\documentclass{article}

     \usepackage[preprint, nonatbib]{neurips_2019}

\usepackage[utf8]{inputenc} 
\usepackage[T1]{fontenc}    
\usepackage{hyperref}       
\usepackage{url}            
\usepackage{booktabs}       
\usepackage{amsfonts}       
\usepackage{amsmath}
\usepackage{microtype}      
\usepackage{graphicx}

\title{Learning depth from monocular video sequences}

\author{%
  Zhenwei Luo \\
  Department of Bioengineering\\
  Rice University \\
  Houston, TX 77030 \\
  \texttt{zl24@rice.edu} \\
}

\begin{document}

\maketitle

\begin{abstract}
Learning single image depth estimation model from monocular video sequence is a very challenging problem. In this paper, we propose a novel training loss which enables us to include more images for supervision during the training process. We propose a simple yet effective model to account the frame to frame pixel motion. We also design a novel network architecture for single image estimation. When combined, our method produces state of the art results for monocular depth estimation on the KITTI dataset in the self-supervised setting.
\end{abstract}

\section{Introduction}
Image based depth estimation is a very important problem in computer vision and has wide applications. An accurate depth map can greatly facilitate further processing such as 3d scene reconstruction and image understanding. Traditionally, estimating depth from images requires special equipment like binocular cameras and LIDAR, or using multi-view stereo methods which requires multiple input images \cite{furukawa_multi-view_2015}. However, humans have superior ability to infer the depth information from a single image. This
ability inspires us to design a general perception machine which acts like human to estimate depth using single image only. With the emergence of deep convolution networks and their unprecedented success in various computer vision related tasks, there has recently been a surge of interest in training deep neural network for monocular depth estimation. One line of work is posing the monocular depth estimation as a supervised learning problem \cite{saxena_learning_2007,
eigen_depth_2014}. Using large
collections of images with ground truth depth data, these methods can train models which produce pixel level depth map. The obvious drawback of these methods is their reliance on the pixel annotated depth map, which costs large efforts for collecting and labeling a large dataset for training deep networks. Another line of work focuses on posing the monocular depth estimation as a self-supervised learning problem \cite{zhou_unsupervised_2017,godard_unsupervised_2017}. These approaches use large
collections of image sequence without any depth information at training time. They are built upon image formation model where the depth map predicted from a single image using deep network serves as an intermediate representation to synthesize images in the sequence. The synthesized images can be supervised by the corresponding images in the data sequence. These self-supervised methods enable us to utilize the abundant low cost monocular video sequences. It’s very promising to obtain highly accurate monocular depth estimation models by training deep neural networks with cheap huge amount of video sequences.

In this paper, we follow the second line of work and propose several improvements to the existing monocular video based depth estimation training framework. Our main contributions are: (1) Designed a novel network architecture for depth estimation. Previous works \cite{godard_digging_2018} demonstrate that the moving object in video sequence can deteriorate the depth estimation network during training. We proposed a novel method to account moving objects in the video sequence. (2) Designed a new
loss function which can leverage more images for supervision in the training process. (3) Trained our network from scratch on the KITTI and Cityscapes datasets by combining the proposed techniques \cite{geiger_are_2012,cordts_cityscapes_2016}. In the self-supervised setting, our models achieve state of the art depth estimation results on the KITTI test set using eigen split \cite{eigen_predicting_2015}.

\section{Related Work}
Training depth estimator using monocular video was first appeared in \cite{zhou_unsupervised_2017, vijayanarasimhan_sfm-net:_2017}. Learning depth estimation model from monocular video is complicated by the unknown camera pose between input frames. \cite{zhou_unsupervised_2017} proposed to use a depth estimation network along with a separate pose estimation network, which is used to predict the camera pose transformation between input frames. Another challenge is moving objects in the recorded scene
may invalidate the rigid view synthesis model for training depth estimation model. \cite{zhou_unsupervised_2017} proposed to downweight loss terms of pixels belong to moving objects using a motion explanation mask, thus reducing the error introduced by moving objects. \cite{vijayanarasimhan_sfm-net:_2017} proposed to explain object motion by a combination of multiple rigid transformations and multiple motion masks. \cite{yin_geonet:_2018} proposed to predict the pixel movement originates from object motion by leveraging the pixel movement
predicted by the camera pose transformation. However, these method shows no improvement to the depth estimation result by including their additional motion explanation terms in the image formation model. The state of the art result prior to this work was presented by \cite{godard_digging_2018}. They proposed several architectural and loss improvements and argued that the depth estimation result can be improved by excluding footage contains moving objects. Our later result will demonstrate that the depth estimation
model can be improved when jointly training with an object motion prediction network using our novel image formation model.

\section{Method}
This section illustrates the basic principle for learning depth predict model from image sequence without any ground truth depth.
The goal of depth estimation can be formulated as learning a function $f$ which predicts a depth map $D$ from a single color image $I$. In the self supervised setting, the depth map $D$ serves as an intermediate representation based on which images for different views can be generated. Real images in video corresponding to those views can then be leveraged to construct a pixelwise photo-consistency loss by comparing with synthesized images. It’s noteworthy that self-supervised monocular depth estimation is an
ill-posed problem as there are many different depth maps for the current view can generate the same image for a novel view given the relative transformation between them. Based on the preceding analysis, learning the depth estimation function $f$ in the self supervised setting can be formulated as minimizing the photo-consistency loss. Given an image $I_t$ for a view $t$, and multiple images $I_{t' }$ for other views $t'$, let $\text{proj}(I_{t' },t\rightarrow t',f(I_t))$ be a image
formation model which generates image for
the view $t$ using $I_{t' }$, its predicted depth map $f(I_t)$, and the relative transformation from view $t$ to $t'$ i.e. $t\rightarrow t'$, we find a model $f$ such that
\begin{equation}
  \min_f {L(I_t,\text{proj}(I_{t^\prime},t\rightarrow t^\prime,f\left(I_t\right)))}
\end{equation}

where $L$ is the photometric loss function.

\subsection{Image Formation Model}
Image formation works by first establishing the relationship between the pixel in the source image and the pixel in the target image. For a pixel with coordinate $[i,j]$, suppose its depth is $d$, and the intrinsic parameter of camera is a 3x3 matrix $K$, the camera coordinate $x$ of the pixel can be expressed as
\begin{equation}
  x = K^{-1} d \begin{pmatrix}
    i \\
    j \\
    1
  \end{pmatrix}.
\end{equation}
Let the transformation matrix which transforms the current camera pose to a new pose be $R$, which is a 4x4 matrix, the camera coordinate of the pixel w.r.t new pose is 
\begin{equation}
  \begin{pmatrix} x' \\
    1
  \end{pmatrix}= R \begin{pmatrix}
    x \\
    1
  \end{pmatrix}.
\end{equation}

The projected coordinate of the pixel can be simply written as
\begin{equation}
  \begin{pmatrix} i' \\
    j' \\
    k'
  \end{pmatrix} = K x'.
\end{equation}

The pixel coordinate in the image taken by the camera with new pose is simply $ [\frac{i'}{k'}, \frac{j'}{k'}] $. We hence show how the pixel coordinate transforms according to camera pose transformation. After projecting the pixel into the image plane of a novel view, we then estimate the pixel value in the original image using pixel values of the image with novel view. \cite{zhou_unsupervised_2017} proposed to use the bilinear sampling algorithm to interpolate the value of the projected pixel with its four
neighboring pixels in the image of novel view, hence generating an image for the original view. This process is commonly known as \textit{inverse warping}. The synthesized image is compared with the original image to compute photometric loss. Following \cite{godard_unsupervised_2017}, denote $I_t$ as frame $t$, we use a combination of $l_1$ and SSIM as our photometric error function \cite{wang_image_2004}, which can be expressed as, 
\begin{equation}
  L(I_t, \bar{I}_t) = \alpha \frac{1 - \text{SSIM}(I_t, \bar{I}_t)}{2} + (1-\alpha)\|I_t - \bar{I}_t\|_1,
\end{equation}
where $\bar{I}_t$ represents the synthesized image. 
The gaussian window size in SSIM is set to 3x3, the standard variance of gaussian window in SSIM is set to 1.5, and  $\alpha$ is set to 0.15. 

The relationship between the pixel coordinate in source image and the pixel coordinate in an image of novel view can be used to estimate the pixel value of the image of novel view, thus allowing us to generate an image for the novel view from the source image. We leverage this observation to design a new image synthetic process where the original image is used to generate images of novel views. Denote pixel coordinate in the image of novel view $t'$ to be interpolated as $[i,j]$, and let
the pixel coordinate in source image be $[i'',j'']$ and its projected pixel coordinate be $[i',j']$, the interpolated pixel value $\bar{I}_{t' } [i,j]$ can be expressed as 
\begin{equation}
  \label{eqn:6}
  \bar{I}_{t' } [i,j]=\sum _{|i-i' |,|j-j' | \le 1} \frac{e^{-\frac{(i-i' )^2+(j-j' )^2}{2}}}{\sum_{|i-i'|,|j-j'| \le 1} e^{-\frac{(i-i')^2+(j-j')^2}{2}}} I_t[i'',j''].
\end{equation}
In other words, the projected pixel is scattered to its neighboring pixels in novel view, and contributes an exponential weighted term to values of its neighboring pixels.
Since our newly designed synthetic process works in reverse direction in contrast to inverse warping, we name it as \textit{warping}. The warped images form new terms in the photometric error function with the corresponding images in video. By introducing this new synthetic process into training, the predict depth map can be supervised by more images in the video sequence. This is helpful to alleviate the ill-posedness of the depth estimation problem.

The synthetic images often present two kinds of systematic errors, i.e., \textit{out of view} and \textit{occlusion}. Out of view occurs in regions of source image which doesn’t present in the target image. For example, when the target view is generated by shifting the source view left, the left border of the original image is shifted out of view and is not presented in the target image, and the right border of the target image contains new contents. In inverse warping, the out of target view pixel will be
projected to pixel at the boundary in target image, thus resulting in large photometric error in those regions. These errors cannot be minimized by improving the depth prediction for source image. Hence, we omit the photometric errors of those out of view pixels. Similar treatments can also be found in \cite{vijayanarasimhan_sfm-net:_2017,mahjourian_unsupervised_2018}. Occlusion is caused by incoherent movements of pixels with different depths when view shifts
\cite{morvan_multi-view_nodate}. If the region around the small depth region is of large depth, the pixel in that region will have very small movement while the pixel with small depth will have very large movement when shifting view. Hence, the pixel with small depth will fall into the region made up of pixels with large depths, which results in large photometric error. 
In inverse warping, we can obtain two images for the same view using its two nearby frames. \cite{godard_digging_2018} proposed to compute the pixelwise minimum from the photometric errors of those two images, which can effectively omit regions with large systematic errors. In this paper, for photometric errors from inverse warping images, we tested both approaches to reduce the effect of systematic errors. The approach in \cite{godard_digging_2018} is referred as minimum photometric error.
However, since warped images are for different views in a batch, it is impossible to compute a pixel minimum. We only use average photometric error for these images.

In the self supervised setting, we use consecutive frames in video for training. Some objects are moving in the world. However, our image formation model stated before assumes that scenes keep static across frames and the transformations among pixels are solely determined by the movements of camera in different frames. With only camera movement in the image formation model, pixels corresponding to the moving objects in source view will be projected into regions 
of other frame where those objects no longer reside. Therefore, this prompts us to introduce extra parameters into the image formation model to account the moving objects. For a pixel $[i,j]$ in the source view, let its relative translation in camera coordinate system be $[t_x,t_y,t_z]$, its camera coordinate w.r.t the source view in a new time frame can be expressed as
\begin{equation}
  x = \begin{pmatrix}
    1 + t_x\\
    1 + t_y\\
    1 + t_z
  \end{pmatrix} \cdot K^{-1} d \begin{pmatrix}
    i \\
    j \\
    1 \\
  \end{pmatrix},
\end{equation}
where $\cdot$ is a elementwise product operator, and $d$ is its depth. We then apply the camera pose transformation to the translated camera coordinate to obtain its new camera coordinate w.r.t target view. The remaining image formation steps are unchanged. The relative translation of each pixel is predicted by an encoder-decoder convolution network whose architecture will detail in later section.

\subsection{Training Loss}

As mentioned before, the depth estimation problem is ill-posed. We address the solution ambiguity of ill-posed problem by enforcing smoothness in the inverse depth maps. Our smoothness restraint is a combination of $l_1$ norms of the first order gradient and the second order gradient, which is of the form
\begin{equation}
  L_s=\sum_{i,j}\beta(|\partial_x d_{ij} |+|\partial_y d_{ij} |)+(1-\beta)(|\partial_{xx} d_{ij} |+|\partial_{yy} d_{ij} |+|\partial_{xy} d_{ij} |+|\partial_{yx} d_{ij} |),
\end{equation}
where $d_{ij}$ is the inverse depth of the pixel $[i,j]$. In this paper, $\beta$ is set to 0.25. For each frame $t$, we use its two neighboring frames for training. The final training loss is a combination of photometric term and smooth term as below,
\begin{equation}
  L = \sum (L'(L(I_t, \bar{I}_t(I_{t+1})), L(I_t, \bar{I}_t(I_{t+1}))) + L(\bar{I}_{t+1}(I_t), I_{t+1}) + L(\bar{I}_{t-1}(I_t), I_{t-1})) + wL_s,
\end{equation}
where $\bar{I}_t(I_{t\pm 1})$ are images synthesized from $I_{t\pm 1}$ using inverse warping,
$\bar{I}_{t\pm1} (I_t )$ represent images synthesized from $I_t$ using equation \ref{eqn:6}, and $L'=L(I_t,\bar{I}_t (I_{t+1} ))+L(I_t,\bar{I}_t (I_{t-1} ))$ when average error is used and $L'=2\min(L(I_t,\bar{I}_t (I_{t+1} )),L(I_t,\bar{I}_t (I_{t-1} ))) $ when minimum error is used. The unknown parameters appear in image formation model are predicted by convolution neural networks. We can train those neural networks by minimizing the loss function. Since the convolution neural networks used in
this study are of encoder-decoder style, we consider further supervising the intermediate layers in the decoder by employing a multiscale loss function as in \cite{zhou_unsupervised_2017}. Following \cite{godard_digging_2018}, the outputs of the intermediate layers are upsampled to the same resolution as input image by bilinear sampler and then involved in the image formation process. This guarantees that our photometric losses are computed between high resolution images. 
\subsection{Network Architecture}
We begin with listing the unknown variables in the image formation model. First of all, the depth map of the source image should be predicted. In the self supervised setting using only monocular video, the relative transformation between camera poses in different frames is unknown. In our new image formation model, we also need to predict the relative translation for each pixel. Each set of unknown variables should be predicted by a neural network to enable end-to-end training. The network
architectures employed in this paper to predict those variables are described below.

Our depth estimator has an U-net style architecture which is common in previous studies \cite{ronneberger_u-net:_2015, zhou_unsupervised_2017,godard_unsupervised_2017} . We use resnet50 as the encoder of our depth estimator \cite{he_identity_2016} and make a few changes to the resnet model; (1) Since the depth estimation is a dense prediction task, we remove the pooling layer after the first convolution layer in original resnet50 architecture to preserve the pixel to pixel correspondence. (2) The stride of
the last block of resnet50 is set to 2 to downsample the resolution of input image by 32. 

\begin{figure}[h]
\centering
\includegraphics[width=0.6\textwidth]{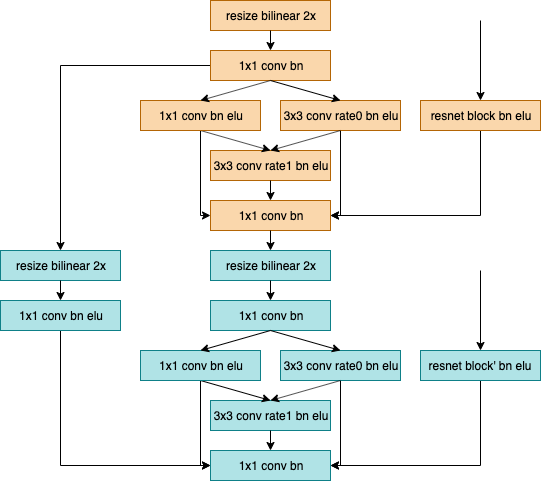}
\caption{The architecture of two consecutive decoder blocks in the depth estimator, where layers of the same block are colored in the same color.}
  \label{fig:figure1}
\end{figure}

The decoder of our depth estimator is inspired
by the astrous spatial pyramid pooling block \cite{chen_deeplab:_2018,zhao_pyramid_2017} in deeplabv3 \cite{chen_rethinking_2017}. We contemplate that depth prediction task needs to leverage multiscale contextual information since objects in scene are of different scales. We hence use dilated convolutions with different dilation parameters to create multiscale convolution filters as in \cite{chen_rethinking_2017,chen_deeplab:_2018}. In addition to the skip connection between decoder and encoder blocks, we
also introduce skip connection to different blocks of decoder to supply multiscale features. The detailed architecture of our decoder is shown in Figure \ref{fig:figure1} and
Table \ref{tab:table1}. Figure \ref{fig:figure1} shows two consecutive decoder blocks. The skip connection between
encoder and decoder refers as the \textit{skipx} layer in Table \ref{tab:table1}. The input of skip connection is denoted as \textit{resnet block} in Figure \ref{fig:figure1}, which are batch normalized and activated by elu before feeding into \textit{skipx} \cite{ioffe_batch_2015, clevert_fast_2015}. \textit{skipx\_1} denotes the skip connection between decoder blocks and is the leftmost branch in Figure \ref{fig:figure1}. The coarser feature from lower block is bilinearly upsampled before entering higher block. A few
implementation details are: (1) The outputs of all
convolution layers except the \textit{dispx} layers in our network are batch normalized before
applying activation. (2) Since the image formation model is invariant to the scale of depth map, we adopt the normalization trick proposed in \cite{wang_learning_2018} which eliminates the scale ambiguity of depth prediction by dividing all predicted inverse depths \textit{dispx} by their mean.

To predict the transformation parameters, we adopt the pose net architecture proposed by \cite{zhou_unsupervised_2017} with some modifications. The pose net takes three consecutive frames as input. The decoder part of pose net was originally designed for outputting explainability mask to weight the loss of each pixel and adopted the transposed convolution layer to upsample the features \cite{zhou_unsupervised_2017}. We use the decoder of pose net to predict the relative translation of each pixel and choose the bilinear sampler which followed by
convolution layer to upsample features in the decoder of pose net. This upsample method can prevent checkerboard artifacts in the transposed convolution upsampled features \cite{odena_deconvolution_2016}. Relu activation function \cite{nair_rectified_2010} is used for all layers of the pose net excepts the transformation output layer which uses no activation. The outputs of pose net are multiplied by 0.01 to keep predictions in a reasonable range. The complete architecture of pose net and the parameter settings for each layer are shown in
Table \ref{tab:table2}. The relative transformation between camera poses are predicted as the channel wise average of the output of layer \textit{pose\_pred} in Table \ref{tab:table2}. \textit{maskx} represents the prediction of relative pixel translations. As in the depth estimation loss, we enforce the smoothness of the relative translations by adding the smoothness term define in equation (8) to training loss function. Since most parts of scene are static, we also enforce the sparseness of
the relative translations by adding the $l_1$ norm of relative translations to training loss function \cite{tibshirani_regression_1996}.

\renewcommand{\tabcolsep}{1mm}
\begin{table}[]
\caption{The complete decoder architecture and parameter settings, where kernel is the size of kernel, rate is the dilation rate of dilated convolution, channel is the number of output channels, res represents the downscaling factor for each layer relative to the input image, activation is the activation function used after convolution, and input corresponds to the input of each layer.}
\label{tab:table1}
\begin{tabular}{lllllll}
  \hline
layer & kernel & rate & channel & res & activation & input \\
\hline
  upnet & 1 & 1 & 1024 & 32 & none & block4 \\
  conv & 1 & 1 & 1024 & 16 & none & upnet \\
  conv\_1 & 1 & 1 & 512 & 16 & elu & conv\\
  conv\_2 & 3 & 2 & 512 & 16 & elu & conv\\
  conv\_3 & 3 & 4 & 512 & 16 & elu & conv\_1, conv\_2 \\
  skip & 1 & 1 & 512 & 16 & elu & block3 \\
  xconcat & 1 & 1 & 512 & 16 & none & conv\_1, conv\_2, conv\_3, skip \\
  \hline
  conv1    & 1 & 1 & 512 & 8 & none & xconcat \\
  conv1\_1 & 1 & 1 & 256 & 8 & elu  & conv1\\
  conv1\_2 & 3 & 2 & 256 & 8 & elu  & conv1\\
  conv1\_3 & 3 & 4 & 256 & 8 & elu  & conv1\_1, conv1\_2 \\
  skip1    & 1 & 1 & 256 & 8 & elu  & block2 \\
  xconcat1 & 1 & 1 & 256 & 8 & none & conv1\_1, conv1\_2, conv1\_3, skip1 \\
  \hline
  conv2    & 1 & 1 & 256 & 4 & none    & xconcat1 \\
  conv2\_1 & 1 & 1 & 128 & 4 & elu     & conv2\\
  conv2\_2 & 3 & 2 & 128 & 4 & elu     & conv2\\
  conv2\_3 & 3 & 4 & 128 & 4 & elu     & conv2\_1, conv1\_2 \\
  skip2    & 1 & 1 & 128 & 4 & elu     & block1 \\
  skip2\_1 & 1 & 1 & 128 & 4 & elu     & conv1 resize bilinear 2x \\
  xconcat2 & 1 & 1 & 128 & 4 & none    & conv2\_1, conv2\_2, conv2\_3, skip2, skip2\_1 \\
  disp2    & 3 & 1 & 1   & 4 & sigmoid & xconcat2\\
  \hline
  conv3    & 1 & 1 & 128 & 2 & none    & xconcat2 \\
  conv3\_1 & 1 & 1 & 64 & 2 & elu     & conv3\\
  conv3\_2 & 3 & 3 & 64 & 2 & elu     & conv3\\
  conv3\_3 & 3 & 6 & 64 & 2 & elu     & conv3\_1, conv3\_2 \\
  skip3    & 1 & 1 & 64 & 2 & elu     & block0 \\
  skip3\_1 & 1 & 1 & 64 & 2 & elu     & conv2 resize bilinear 2x \\
  xconcat3 & 1 & 1 & 64 & 2 & none    & conv3\_1, conv3\_2, conv3\_3, skip3, skip3\_1 \\
  disp3    & 3 & 1 & 1  & 2 & sigmoid & xconcat3\\
  \hline
  conv4    & 1 & 1 & 64 & 1 & none    & xconcat3 \\
  conv4\_1 & 1 & 1 & 32 & 1 & elu     & conv4\\
  conv4\_2 & 3 & 3 & 32 & 1 & elu     & conv4\\
  conv4\_3 & 3 & 6 & 32 & 1 & elu     & conv4\_1, conv4\_2 \\
  conv4\_4 & 1 & 12 & 32 & 1 & elu     & conv4\_2,conv4\_3 \\
  skip4\_1 & 1 & 1 & 32 & 1 & elu     & conv3 resize bilinear 2x \\
  xconcat4 & 1 & 1 & 32 & 1 & none    & conv4\_1, conv4\_2, conv4\_3, skip4, skip4\_1 \\
  disp4    & 3 & 1 & 1  & 1 & sigmoid & xconcat4\\
  \hline

\end{tabular}
\end{table}

\begin{table}[]
  \caption{The complete architecture of pose net and parameter settings. Kernel, channel, res, activation and input represent the same parameters as in Table \ref{tab:table1}. Stride is the stride step of convolution}
\label{tab:table2}
\begin{tabular}{lllllll}
  \hline
layer & kernel & stride & channel & res & activation & input \\
\hline
  cnv1	  &7	&2	&16	  &2	&relu	&image\\
cnv2		&5		&2	&32	  &4	&relu	&cnv1\\
cnv3		&3		&2	&64	  &8	&relu	&cnv2\\
cnv4		&3		&2	&128	&16	&relu	&cnv3\\
cnv5		&3		&2	&256	&32	&relu	&cnv4\\
\hline
  cnv6  &3    &2  &256  &64 &relu &cnv5\\
  cnv7  &3    &2  &256  &128 &relu &cnv6\\
pose\_pred  &1    &1  &12  &128 &none &cnv7\\
\hline
  upcnv5	&3	&1	&256	&16	&relu	&cnv5 resize bilinear 2x\\
upcnv4		&3	&1	&128	&8	&relu	&upcnv5 resize bilinear 2x\\
\hline
  upcnv3  &3  &1  &64   &4  &relu &upcnv4 resize bilinear 2x\\
  mask3   &3  &1  &6    &4  &none &upcnv3\\
  \hline
  upcnv2  &3  &1  &32   &2  &relu &upcnv3 resize bilinear 2x\\
  mask2   &3  &1  &6    &2  &none &upcnv2\\
  \hline
  upcnv1  &3  &1  &16   &1  &relu &upcnv2 resize bilinear 2x\\
  mask1   &3  &1  &6    &1  &none &upcnv1\\

\hline
\end{tabular}
\end{table}

\subsection{Implementation Details}
We implemented our neural networks in tensorflow\cite{abadi_tensorflow:_2016}. The weights of smoothness term or sparseness term in training loss were resolution dependent and set to $\frac{0.01}{2^s}$ , where $s$ was the downscaling factor for the output on which the regularization term was applied relative to the input image. We employed the nadam optimizer to minimize the training loss\cite{dozat_incorporating_2016}. The learning rate was set to $10^{-4}$ throughout the training process, the momentum related parameters $\beta_1$ and $\beta_2$ were set to 0.9 and 0.999,
respectively. The $\varepsilon$ value was $10^{-8}$. The batch size was 8. We resized the images of the KITTI dataset to 128x418 pixels. For Cityscapes dataset, we first cropped out the bottom 25\% of image that contains car logo, and then resized the cropped image to 128x418 pixels. The input images were subject to data augmentation process before feeding into neural networks. All rgb values of input image were first scaled by a random factor between 5/6 and 1.2, and the values of each rgb
channel were multiplied with an independent random factor between 0.8 and 1.2. We also adjusted the contrast of image by a random factor between 0.5 and 1.5. The left and right part of the image was swapped with a probability 0.5. Finally, our models were exponential moving averages of the trained parameters. The decay parameter of exponential moving average was set to 0.9997. The models were deposited after every 5100 iterations. Our neural networks were trained from scratch with default
initialization method. The implementation of our method is located at https://github.com/alncat/SFM.

\section{Results}
We evaluated the depth prediction result on KITTI 2015 using the eigen split \cite{eigen_predicting_2015}. The maximum depth of the dataset involved in evaluation is capped at 80 meters as in previous works. The predicted depth map for evaluation is the average of all predicted normalized depth maps of the decoder. The low resolution depth maps are bilinearly upsampled to the same resolution as input
image. Before evaluation, the predicted depth map is put on the same scale as the ground truth depth map by comparing its median with the median of the ground truth depth map \cite{zhou_unsupervised_2017}. As
noted in \cite{godard_digging_2018}, the original evaluation code from \cite{zhou_unsupervised_2017}
used by most subsequent works adopted an incorrect flag when generating the ground truth depth map. The ground truth depth map generated by the incorrect evaluation code was computed w.r.t the LIDAR instead of the cameras. For consistency, we presented our results evaluated by incorrect code. In all metrics presented in tables, smaller AbsRel, SqRel, RMSE, and RMSELog values are better, while larger $\delta < x$ values are better. The best model trained with average loss and CK was found at
 iteration 907800, and the best model trained with minimum loss and CK was found at iteration 576300. For models trained with K, the best model using average loss was found at iteration 515100, and the best model using minimum loss was found at iteration 357000. As it is shown in Table \ref{tab:table3}, our result using the
model trained with KITTI and Cityscapes datasets outperforms prior best result in all metrics. Especially in criteria that are sensitive to large depth
errors i.e. square relative error and root mean square error, the improvements are more noticeable. Our model trained solely on KITTI also has comparable performance w.r.t the prior best result, Godard HR, which is trained using images with higher resolution and pretrained encoder. It even outperforms Godard HR on metrics like AbsRel and RMSE. For completeness, we also presented our results evaluated by correct code in Table \ref{tab:table4}. It is easy to see that our
results are greatly improved over Godard HR.
\begin{table}[]
  \caption{Results for different models which are trained with monocular video sequences on KITTI 2015 using Eigen’s split \cite{eigen_predicting_2015}. K represents model trained with KITTI 2015\cite{geiger_are_2012}. CK represents model trained using a combination of Cityscapes \cite{cordts_cityscapes_2016} and KITTI 2015. HR represents model trained with images of size 192x640. Avg means the model is trained with the average photometric error. Min means the model is trained with the pixel minimum photometric error. Best results are bolded.}
  \label{tab:table3}
\begin{tabular}{lllllllll}
  \hline
Method & Dataset & AbsRel & SqRel & RMSE & RMSELog & $\delta < 1.25$ & $\delta < 1.25^2$ & $\delta<1.25^3$ \\
\hline
  Zhou\cite{zhou_unsupervised_2017}	      & K	  &0.183	&1.595	&6.709	&0.270	&0.734	&0.902	&0.959\\ 
  Yang\cite{yang_unsupervised_2017}	      & K	  &0.182	&1.481	&6.501	&0.267	&0.725	&0.906	&0.963\\ 
  GeoNet\cite{yin_geonet:_2018}	    & K	  &0.155	&1.296	&5.857	&0.233	&0.793	&0.931	&0.973\\  
  DDVO\cite{wang_learning_2018}	      & K	  &0.151	&1.257	&5.583	&0.228	&0.810	&0.936	&0.974\\  
  Godard\cite{godard_digging_2018}	    & K	  &0.133	&1.158	&5.370	&0.208	&0.841	&0.949	&0.978\\  
  Godard HR\cite{godard_digging_2018}	  & K	  &0.129	&1.112	&5.180	&0.205	&0.851	&0.952	&0.978\\ 
  Godard\cite{godard_digging_2018}	    & CK	&0.138	&1.430	&5.609	&0.215	&0.843	&0.948	&0.975\\ 
Ours Avg	  & CK	&\textbf{0.122}	&0.871	&5.012	&0.204	&\textbf{0.855}	&0.953	&0.979\\ 
  Ours Min	  & CK	&\textbf{0.122}	&\textbf{0.820}	&\textbf{4.856}	&\textbf{0.198}	&\textbf{0.855}	&\textbf{0.955}	&\textbf{0.981}\\ 
Ours Avg	  & K	  &0.124	&0.878	&5.189	&0.206	&0.846	&0.950	&0.979\\ 
Ours Min	  & K	  &0.124	&0.881	&5.157	&0.205	&0.845	&0.950	&0.979 \\
\hline
\end{tabular}
\end{table}

\begin{table}[]
  \caption{Results for different models evaluated by revised code on KITTI 2015 using Eigen’s split. K, CK, HR, Avg and Min refer to the same settings as in Table \ref{tab:table3}. Best results are bolded.}
  \label{tab:table4}
\begin{tabular}{lllllllll}
  \hline
Method & Dataset & AbsRel & SqRel & RMSE & RMSELog & $\delta < 1.25$ & $\delta < 1.25^2$ & $\delta<1.25^3$ \\
\hline
  Godard \cite{godard_digging_2018}	  &K	&0.137	&1.153	&5.353	&0.212	&0.836	&0.947	&0.978\\ 
  Godard HR	\cite{godard_digging_2018}  &K	&0.133	&1.111	&5.182	&0.209	&0.845	&0.950	&0.977\\
  Ours Avg	  &CK	&\textbf{0.124}	&0.876	&5.040	&0.206	&0.849	&0.951	&0.979\\
  Ours Min	  &CK	&\textbf{0.124}	&\textbf{0.825}	&\textbf{4.882}	&\textbf{0.202}	&\textbf{0.850}	&\textbf{0.952}	&\textbf{0.980}\\
Ours Avg	  &K	&0.128	&0.894	&5.234	&0.212	&0.838	&0.948	&0.978\\
Ours Min	  &K	&0.128	&0.894	&5.198	&0.210	&0.838	&0.947	&0.978\\
\hline
\end{tabular}
\end{table}
\section{Conclusion}
We demonstrated that our method can achieve state-of-the-art depth estimation result in the self supervised setting with architectural and loss innovations. We proposed estimating the relative translations of each pixel for each frame to account the moving objects in scenes. The prior state-of-the-art method \cite{godard_digging_2018} argued that the object motion in monocular video sequence can undermine the training of depth estimation network and demonstrated that their depth estimator had
worse performances when trained with the dataset with large portion of moving objects i.e. Cityscapes. However, by jointly training with our newly design relative translation prediction network, our depth estimation network achieved better performance on the combination of Cityscapes and KITTI dataset. This proves the effectiveness our object motion explanation procedure and shows that it enables us to leverage more dynamic datasets such as Cityscapes to train monocular depth estimation network.
Depth estimation is a well known ill-posed problem. By combining the inverse warping with our newly designed warping process, we increased the number of images for supervising the depth of a single image, thus improving the quality of depth estimation result. A future direction to evolve our self supervised depth estimation method might be improving the architecture of pose and relative translation estimation network since we employ a simple design in this work. We may also test
training depth estimation network with mixture forms of data, i.e., stereo and monocular video, using our new loss.



\small
\bibliography{ref}
\bibliographystyle{plain}
\end{document}